\documentclass[10pt,twocolumn,letterpaper]{article}

\usepackage{iccv}              

\definecolor{iccvblue}{rgb}{0.21,0.49,0.74}
\usepackage[pagebackref,breaklinks,colorlinks,allcolors=iccvblue]{hyperref}
\usepackage{amsthm,amsmath,amssymb}
\usepackage{multirow}
\usepackage{mathrsfs}
\usepackage{xcolor}
\usepackage{color, colortbl}
\usepackage{pifont}

\usepackage{marvosym} 
\usepackage{fontawesome} 

\def\paperID{7491} 
\def\confName{ICCV}
\def\confYear{2025}

\title{FreqPDE: Rethinking Positional Depth Embedding for Multi-View 3D Object Detection Transformers}

\author{Haisheng Su$^{1,3\spadesuit}$ \quad
Junjie Zhang$^{2,3\spadesuit\heartsuit}$ \quad
Feixiang Song$^{3}$ \quad 
Sanping Zhou$^{2}$ \\
Wei Wu$^{3}$ \quad
Nanning Zheng$^{2}$$^{\textrm{\Letter}}$ \quad
Junchi Yan$^{1}$$^{\textrm{\Letter}}$  \\
$^{1}$Shanghai Jiao Tong University, $^{2}$Xi'an Jiaotong University, $^{3}$SenseAuto Research \\
{\tt\small \{suhaisheng,yanjunchi\}@sjtu.edu.cn, hooz1009@stu.xjtu.edu.cn, nnzheng@mail.xjtu.edu.cn} \\
\vspace{-0.6cm}
}

\begin{document}
\maketitle

\let\thefootnote\relax\footnotetext{$\spadesuit$ Equal Contribution.\quad \quad $\textrm{\Letter}$ Corresponding Authors.}
\let\thefootnote\relax\footnotetext{$\heartsuit$ Work done during an internship at SenseAuto Research.}

\begin{abstract}
Detecting 3D objects accurately from multi-view 2D images is a challenging yet essential task in the field of autonomous driving. Current methods resort to integrating depth prediction to recover the spatial information for object query decoding, which necessitates explicit supervision from LiDAR points during the training phase. However, the predicted depth quality is still unsatisfactory such as depth discontinuity of object boundaries and indistinction of small objects, which are mainly caused by the sparse supervision of projected points and the use of high-level image features for depth prediction. Besides, cross-view consistency and scale invariance are also overlooked in previous methods. In this paper, we introduce Frequency-aware Positional Depth Embedding (FreqPDE) to equip 2D image features with spatial information for 3D detection transformer decoder, which can be obtained through three main modules. Specifically, the Frequency-aware Spatial Pyramid Encoder (FSPE) constructs a feature pyramid by combining high-frequency edge clues and low-frequency semantics from different levels respectively. Then the Cross-view Scale-invariant Depth Predictor (CSDP) estimates the pixel-level depth distribution with cross-view and efficient channel attention mechanism. Finally, the Positional Depth Encoder (PDE) combines the 2D image features and 3D position embeddings to generate the 3D depth-aware features for query decoding. Additionally, hybrid depth supervision is adopted for complementary depth learning from both metric and distribution aspects. Extensive experiments conducted on the nuScenes dataset demonstrate the effectiveness and superiority of our proposed method.


\end{abstract}    
\vspace{-0.3cm}
\section{Introduction}
\label{sec:intro}

\begin{figure}[t]
\centering
\includegraphics[width=0.9\columnwidth]{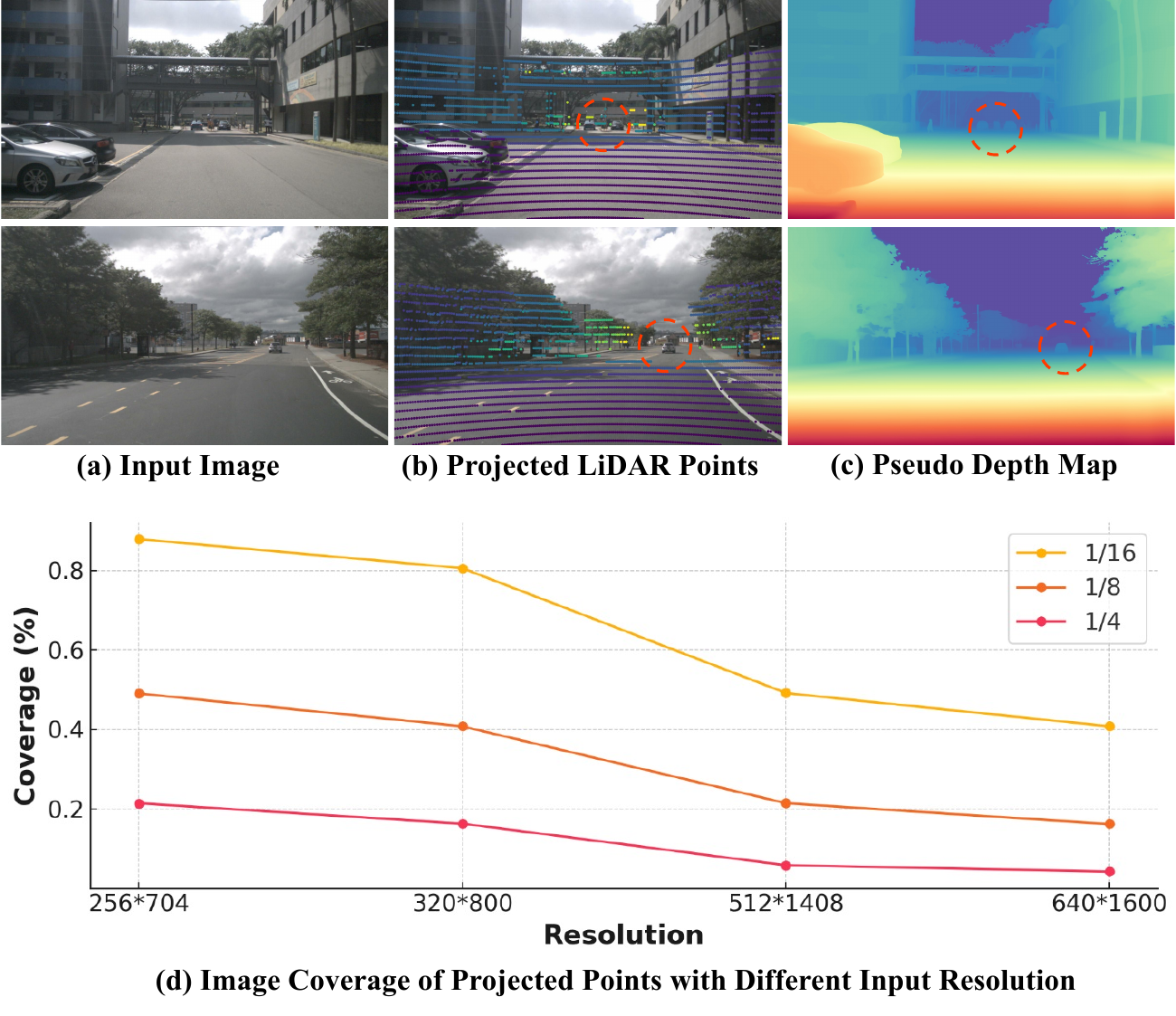}   
\vspace{-0.3cm}
\caption{Illustration of motivation. (a) Input image with a fixed resolution. (b) Missing depth supervision of distant vehicles \textcolor{red}{(red circle)} with sparse projected LiDAR points. (c) Pseudo depth map complements implicit distribution priors additionally. (d) Coverage comparisons of projected points with different input resolution and downsampling factors.}
\label{fig::overview}
\vspace{-0.5cm}
\end{figure}


In recent years, great advancements in perception technology~\cite{weng2019monocular,reading2021categorical,wang2024multi,lan2022arm3d} have been witnessed in autonomous driving. Compared to LiDAR-based 3D perception applications, the camera-based paradigm has drawn increasing attention from both industrial and academic researchers owing to the controllable cost. However, recovering the 3D spatial information from multi-view 2D images is an ill-posed problem, and current methods usually integrate the depth prediction task for introducing additional supervision.

Literally, recent works attempt to exploit depth information in various ways. Pseudo-LiDAR~\cite{weng2019monocular} estimates dense depth map and generates the pseudo 3D lidar points using the camera extrinsic which can be processed with the off-the-shelf LiDAR-based 3D detector. CaDDN~\cite{reading2021categorical} and BEVDet series~\cite{huang2021bevdet, huang2022bevdet4d, huang2022bevpoolv2} predict the pixel-wise probabilistic depth distribution and then project the 2D image features to the 3D space following the Lift-Splat-Shoot (LSS)~\cite{philion2020lift} paradigm. BEVDepth~\cite{li2023bevdepth} introduces the explicit supervision of the depth branch with the projected lidar points as shown in Fig.~\ref{fig::overview} (b), which increases the depth prediction accuracy and detection performance significantly. 3DPPE~\cite{shu20233dppe} proposes the 3D point positional encoding to generate position-aware 3D features and performs the position embedding transformation following~\cite{liu2022petr}.
Undoubtedly, the quality of predicted depth determines the performance upper limit of depth-based detectors. However, the predicted depth quality remains unsatisfactory in three main aspects: (1) only high-level image features are used for depth prediction, leading to depth discontinuity of object boundaries and indistinction of small objects, owing to the loss of local details during the downsampling process. (2) Sparse supervision of projected lidar points also accounts for incomplete depth learning. As shown in Fig.~\ref{fig::overview} (d), the image coverage of projected lidar points decreases significantly with the increase of input image resolution. (3) Moreover, the independent visual feature extraction and depth prediction inevitably neglect the cross-view consistency and scale invariance.

To this end, we propose FreqPDE, a frequency-aware positional depth embedding, to equip 2D visual features with high-quality spatial information for the 3D detection transformer decoder. Specifically, three main modules are designed to relieve the above issues accordingly. First, the Frequency-aware Spatial Pyramid Encoder (FSPE) handles the extracted visual features to construct a multi-scale feature pyramid through combining high-frequency local details and low-frequency global semantics from different levels respectively. Then the Cross-view Scale-invariant Depth Predictor (CSDP) is designed to perform hierarchy depth prediction with cross-view attention and efficient channel attention, ensuring cross-view consistency and scale invariance. Besides, hybrid depth supervision is introduced to facilitate complementary depth learning from both explicit metric and implicit distribution aspects, with the help of sparse lidar maps and dense pseudo depth maps as shown in Fig.~\ref{fig::overview} (b) and (c) respectively. Finally, the Positional Depth Encoder (PDE) combines the 2D image features and multi-scale positional embeddings to generate 3D depth-aware features, which are adapted for object query decoding with a detection transformer. In sum, the main contributions of our work are three folds:

\begin{itemize}
  \item We propose a Frequency-aware Positional Depth Encoder for high-quality depth prediction, named \textbf{FreqPDE}, which is proven to be effective for improving 3D detection transformers from multi-view perspectives.

  \item We introduce a plug-and-play depth predictor to perform hierarchy depth prediction upon the constructed feature pyramid with frequency enhancement. Besides, cross-view attention and camera-aware channel attention are conducted consecutively to ensure cross-view consistency and scale invariance under the hybrid supervision of both explicit metric and implicit distribution levels.

  \item Extensive experiments conducted on nuScenes~\cite{caesar2020nuscenes} demonstrate the prominent effectiveness of our proposed FreqPDE, revealing the great potential of the high-quality positional embedding for 3D detection transformers.
\end{itemize}

\section{Related Work}
\label{sec:related_work}

\subsection{Multi-view 3D Object Detection}

The advancement of autonomous systems underscores the critical role of surround-view 3D object detection for safety and comfort~\cite{liu2023sparsebev, li2022bevformer,wang2022detr3d,huang2023fast,li2023bevdepth,huang2022bevdet4d}. The BEVDet series~\cite{huang2021bevdet, huang2022bevdet4d, huang2022bevpoolv2, li2023bevdepth, huang2023fast} constructs Bird-Eye-View (BEV) features with 2D to 3D view transformation. DETR3D~\cite{wang2022detr3d} employs transformers to implicitly convert image features and object queries from 2D to 3D, enabling direct 3D object detection following the DETR~\cite{carion2020end} paradigm. Polar DETR~\cite{chen2022polar} enhances feature interaction by reformulating positional parameterization. BEVFormer~\cite{li2022bevformer} integrates both spatial and temporal information through interaction with spatial and temporal spaces via predefined grid-shaped BEV queries. SparseBEV~\cite{liu2023sparsebev} introduces scale-adaptive self-attention and adaptive spatio-temporal sampling to boost performance. StreamPETR~\cite{wang2023exploring} achieves performance comparable to LiDAR-based methods by employing object-centric temporal modeling and maintaining a memory queue for storing historical object queries.


\begin{figure*}[t]
\centering
\includegraphics[width=0.9\textwidth]{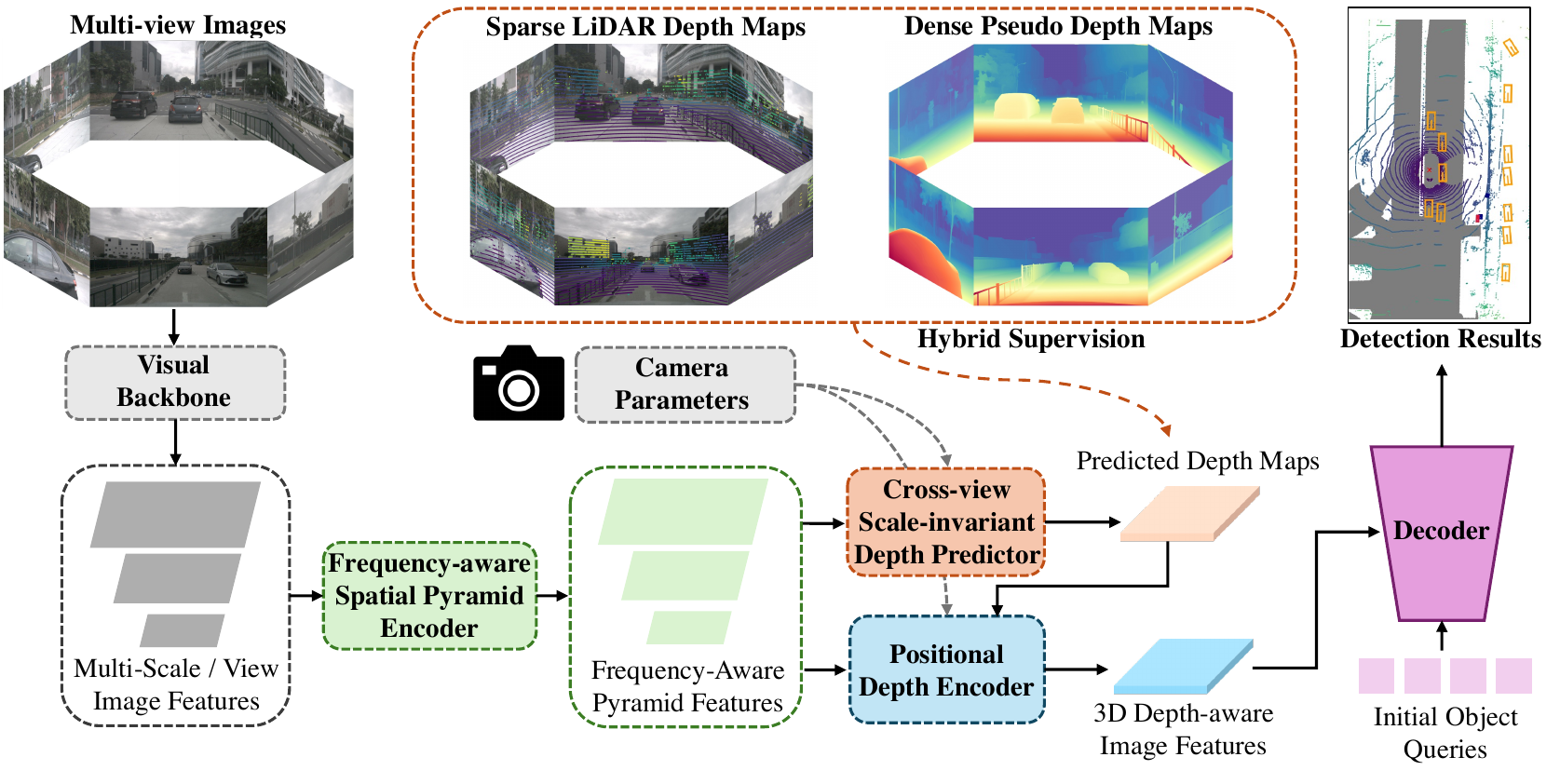}    
\vspace{-0.5cm}
\caption{Overview of our proposed FreqPDE framework. FreqPDE first extracts multi-scale features from multi-view images using an off-the-shelf visual encoder. Then the FSPE module constructs a feature pyramid through combining high-frequency edge clues and low-frequency semantics from low-level and high-level image features respectively. The CSDP module estimates multi-scale depth maps with cross-view attention based on the frequency-enhanced features. Finally, the PDE module encodes both semantic features and geometric depth embedding to generate 3D depth-aware features for object query decoding. Hybrid depth supervision is introduced to facilitate complementary depth learning.}
\label{fig_pipeline}
\end{figure*}

\subsection{Frequency Domain Learning}
In recent years, frequency domain analysis methods, central to signal processing, have seen significant advancements in deep learning. Works~\cite{rahaman2019spectral,xu2021deep} discovered deep learning models tend to prioritize learning low-frequency information. Richard Zhang~\cite{zhang2019making} and Zou \textit{et al.}~\cite{zou2023delving} applied classical anti-aliasing from signal processing to deep learning, enhancing the shift-equivariance of the model. FLCP~\cite{grabinski2022frequencylowcut} proposed an alias-free down-sampling method to address the reduction in model robustness caused by frequency aliasing. Chen \textit{et al.}~\cite{chen2023instance} improves instance segmentation accuracy in low-light environments by suppressing high-frequency noise in image features. SSAH~\cite{luo2022frequency} suggests that perturbing high-frequency noise causes intra-category similarity inconsistency. DFSA~\cite{magid2021dynamic} argues that the absence of high-frequency signals leads to boundary displacement. FreqFusion~\cite{chen2024frequency} employs both high-pass and low-pass filters to retain both high-frequency and low-frequency information, thereby addressing intra-category similarity inconsistency and boundary displacement.


\section{Our Approach}

\subsection{Overview Architecture}
The overall framework of the proposed FreqPDE is illustrated in Fig.~\ref{fig_pipeline}, which aims to improve the 3D detection performance from multi-view perspectives by introducing frequency-aware positional depth embedding. Specifically, FreqPDE mainly consists of three modules, namely FSPE, CSDP and PDE. First, the FSPE module is designed to encode multi-scale spatial features by combining high-frequency edge clues and low-frequency semantics from different levels in a pyramidal structure, respectively. Then the CSDP module conducts hierarchy depth prediction with cross-view consistency and scale invariance. Finally, the PDE module encodes the positional depth values to obtain depth-aware 3D features through element-wise addition of image features and 3D PE, which are used to decode object queries for 3D detection.

\subsection{Frequency-aware Spatial Pyramid Encoder}
Previous detection frameworks typically implement a Feature Pyramid Network (FPN)~\cite{lin2017feature} to reconstruct a set of multi-scale features with abundant semantic information in a top-down pathway. However, Sapa~\cite{lu2022sapa} discovered that outputs from the simple and direct interpolation used by FPN often lean towards excessive smoothness, resulting in boundary displacement~\cite{chen2024frequency} issue. Meanwhile, there exists the spectral bias in DCNNs that these networks prioritize learning the low-frequency modes~\cite{rahaman2019spectral}, which leads to the lack of high-frequency information in high-level features, affecting the depth prediction at the edges of 3D objects and localization accuracy.

\begin{figure}[t]
\includegraphics[width=\columnwidth]{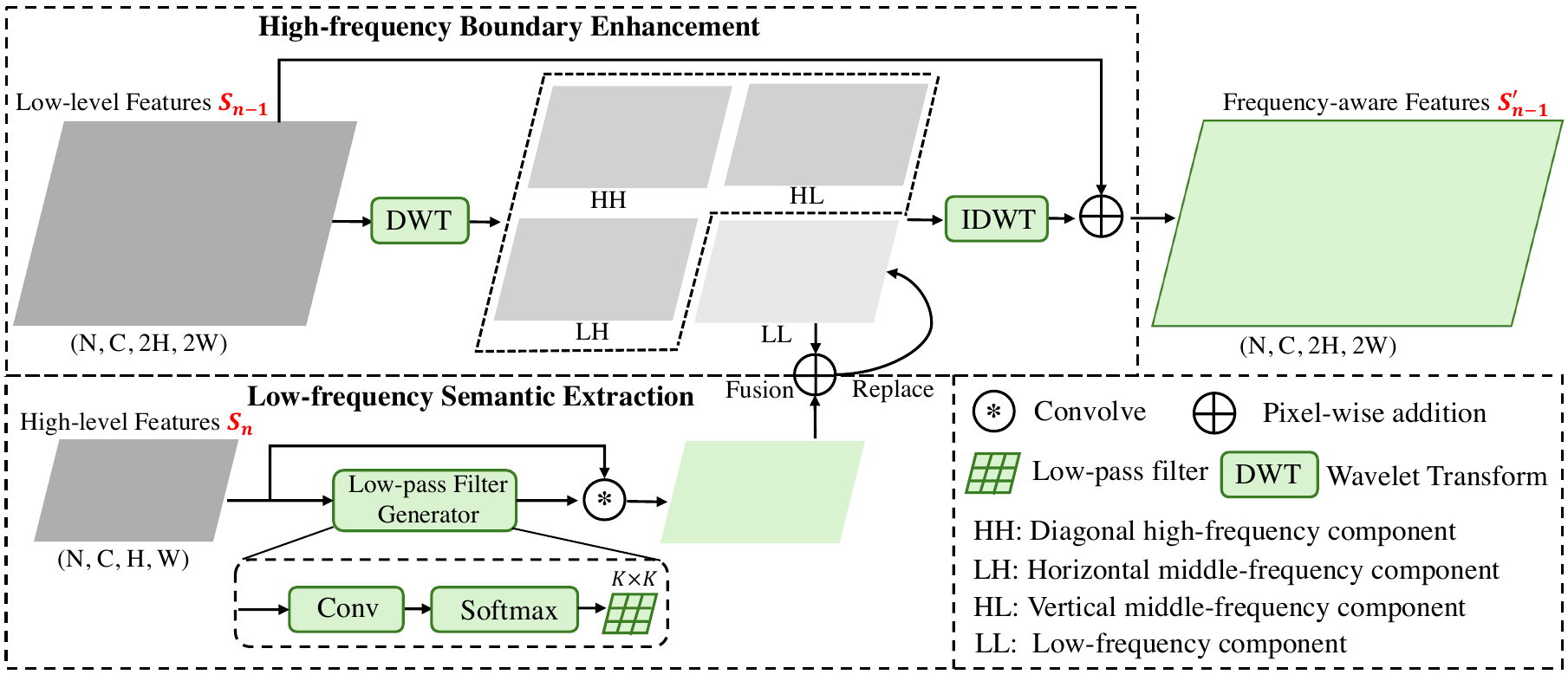}
\vspace{-0.5cm}
\caption{Illustration of FSPE module. After encoding multi-scale features, low-frequency semantics are extracted from the high-level features through adaptive low-pass filters. Then the Discrete Wavelet Transform (DWT) is conducted on the low-level features to decompose four sub-components of different frequencies. The low-frequency component of low-level features is combined with the extracted high-level semantics for boundary enhancement.}
\label{fig_fafpn}
\end{figure}

To overcome the aforementioned problems, we design the FSPE module which is composed of several blocks and each block consists of high-frequency boundary enhancement and low-frequency semantic extraction modules.  Taking one layer of FSPE as an example, with high-level features $S_n\in {\rm R}^{C \times H\times W}$ and low-level features $S_{n-1}\in {\rm R}^{C \times 2H\times 2W}$ as input, the previous method~\cite{lin2017feature} directly interpolates and upsamples $S_n$ which will be added to $S_{n-1}$ directly, this process can be formulated as:
\begin{equation}
\label{eq1}
S_{n-1}' = f_{\rm up}(S_n) + S_{n-1},
\end{equation}
where $f_{\rm up}$ indicates upsampling methods, $S_{n-1}'$ denotes the fused low-level feature which is also the high-level feature input of the next block in the top-down multi-scale feature-building pathway. Instead, with the same inputs, our method can be formulated as:
\begin{equation}
\label{eq2}
S_{n-1}' = f_{\rm iDWT}(f_{\rm lf}(S_{n-1}), f_{\rm hf}(S_n)) + S_{n-1},
\end{equation}
where $f_{\rm lf}$ and $f_{\rm hf}$ denotes the low-frequency semantic extraction and high-frequency boundary enhancement modules, which can utilize high and low-frequency information adaptively, $f_{\rm iDWT}$ is the inverse discrete wavelet transform (iDWT), as illustrated in Fig.~\ref{fig_fafpn}.



\noindent
\textbf{Low-frequency Semantic Extraction.}
To eliminate the smooth and inaccurate boundaries caused by simple interpolation from high-level features during the upsampling process, we utilize low-pass filters which are generated dynamically~\cite{zou2023delving} to effectively extract the global semantics of high-level features as shown in Fig.~\ref{fig_fafpn}. In particular, we take the high-level features $S_{n}$ as input to predict spatial-variant low-pass filters with a learnable method, which comprises a $3\times 3$ convolutional layer and a Softmax layer, this can be represented as:
\begin{equation}
\label{eq3}
W={\rm{Softmax}}({\rm{Conv}}_{3\times 3} (S_n)),
\end{equation}
where $W\in {\rm R}^{K^2 \times H\times W}$, $K$ is the kernel size of predicted low-pass filters, each of the $K^2$ channels indicates a weight at one of the $K\times K$ locations within the filters, and the kernel-wise softmax is utilized to constrain these weights to be positive and sum to one. Then we apply $W$ to input $S_n$ to generate context-aware high-level feature as follows:
\begin{equation}
\label{eq4}
\overline{S_n^{i,j}} = \sum_{p,q\in \Omega }W_{p,q}^{i,j}\cdot  S_n^{i+p,j+q},
\end{equation}
where $\overline{S_n^{i,j}}$ is the output feature at location $(i,j)$ and $\Omega$ is the $K\times K$ region surrounding positions $(i,j)$, in which we apply the pixel-wise product and sum together, new feature $\overline{S_n}$ will be used for reconstructing low-level feature afterward.

\noindent
\textbf{High-frequency Boundary Enhancement.}
Inspired by some lossless wavelet methods~\cite{graps1995introduction}, we introduce DWT to preserve high-frequency information from low-level features of higher resolution. Given input low level feature $S_{n-1}$, DWT is performed first to split $S_{n-1}$ into 4 sub-components($LL$, $LH$, $HL$, $HH$) as described in Fig.~\ref{fig_fafpn}, where all the components have identical shapes with $\overline{S_n}$. Although the low-frequency components $LL$ possess certain global semantic information, they lack the response to salient features due to the average pooling operation~\cite{yl2010theoretical}. Therefore, we add high-level feature $\overline{S_n}$ with $LL$ to construct a fresh component $LL'$ which can be performed with iDWT operation. This method not only avoids the boundary displacement issue caused by upsampling methods, but also losslessly retains the high-frequency details corresponding to the larger size. Finally, we add the original low-level feature $S_{n-1}$ to the new feature residually to obtain the frequency-aware feature $S_{n-1}'$, this can be formulated as:
\begin{equation}
\label{eq5}
S_{n-1}' = f_{\textbf{iDWT}}(LL',(LH,HL,HH))+S_{n-1}
\end{equation}

\begin{figure*}[t]
\centering
\includegraphics[width=0.95\textwidth]{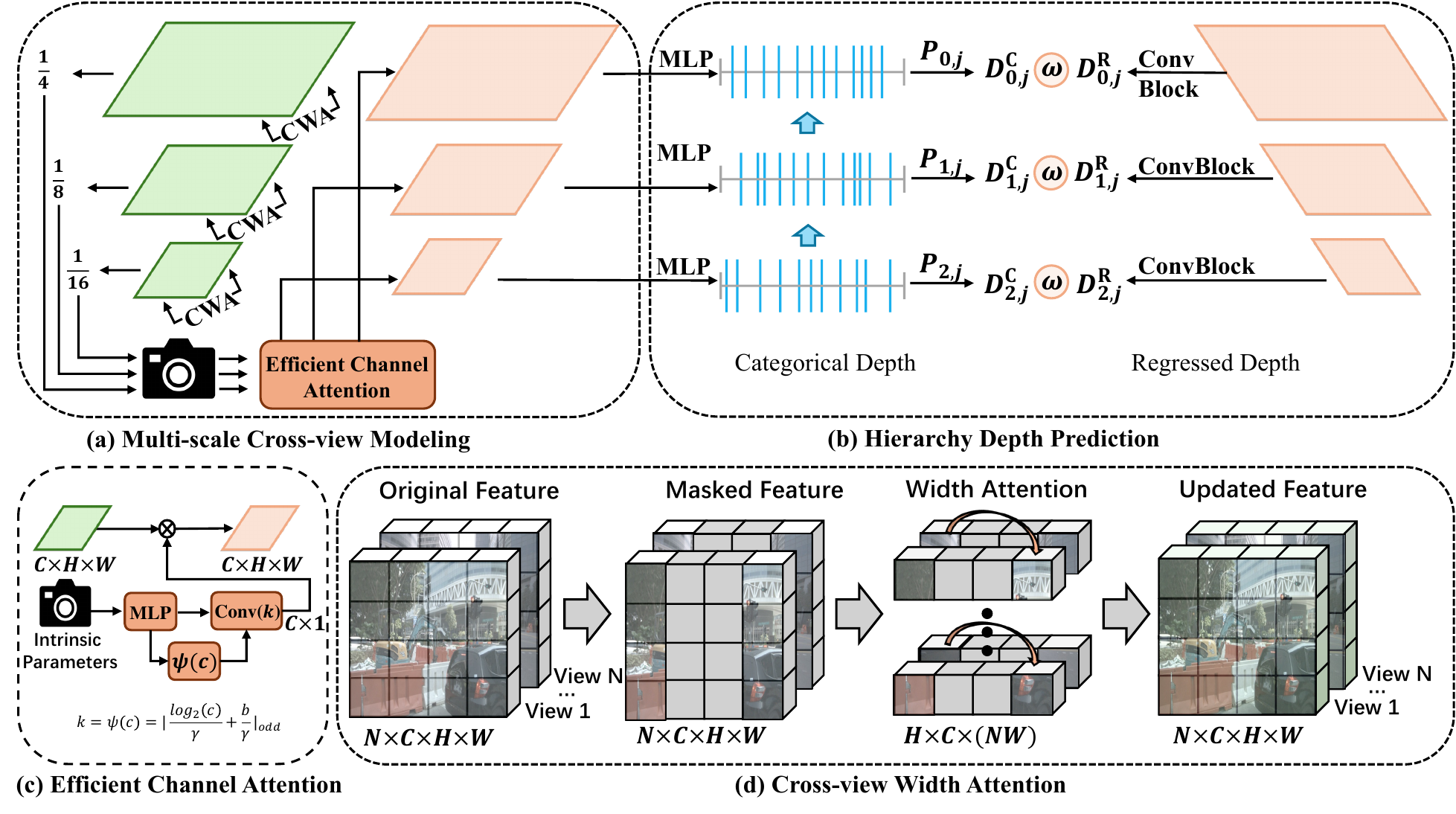}
\vspace{-0.4cm}
\caption{Detailed structure of CSDP module. (a) illustrates the process of visual feature re-weighting with efficient channel attention and cross-view attention from different levels to achieve cross-view consistency and scale invariance respectively. (b) describes the details of hierarchy depth prediction, where both categorical and regressed depth are predicted accordingly, which are fused with a hyper-parameter $\omega$ to generate the final depth map. (c) and (d) show the process of efficient channel attention and cross-view attention respectively.}
\label{fig_depth_head}
\vspace{-0.2cm}
\end{figure*}
 
\subsection{Cross-view Scale-invariant Depth Prediction}
Previous works~\cite{tian2020fcos, wang2021fcos3d} have demonstrated that features at different levels significantly enhance the detection performance for objects of various sizes, which inspires us to build a multi-scale depth head to predict a more accurate depth map. Specifically, in contrast to the other depth prediction framework, we exploit multi-scale output features of FSPE to predict the depth maps of the corresponding size with hybrid-depth module~\cite{shu20233dppe}, the structure of our depth head is shown in Fig.~\ref{fig_depth_head}.

\noindent
\textbf{Cross-view Modeling.}
With multi-view image feature input, it is crucial to ensure consistency across different perspectives during the feature extraction and depth prediction process. Therefore, we introduce a novel Cross-view Width Attention (CWA) block to facilitate interaction in the overlapping regions between the image features of adjacent views. As illustrated in Fig.~\ref{fig_depth_head} (d), we first mask the feature with a fixed ratio $\mu$ to ignore the influence of non-overlapping regions. Then we reshape the multi-view features into groups according to their row indices and the width-attention~\cite{liu2023geomim} mechanism is performed to ensure width-wise features only interact with the others belonging to the same row. Moreover, to avoid increasing the model parameters and computation significantly, we only insert one CWA block to the features at each layer. 

\noindent
\textbf{Scale-invariant Learning.} 
To associate depth prediction with the camera intrinsics, we scale up the dimension of camera intrinsics to the features with an MLP layer, then we introduce ECANet~\cite{wang2020eca} to re-weight the input features $F_{i,j} \in {\rm R}^{C \times H_i\times W_i}$, this process can be written in:
\begin{equation}
\label{eq5}
F_{i,j}'={\rm ECA}(F_{i,j} |{\rm MLP}({\zeta}_{i}K_{j})),
\end{equation}
where $i=1,\cdots,l$, indicate the level of input features, $j=1,\cdots,6$, indicate the index of the cameras, $K_{j}$ is the corresponding camera intrisics and ${\zeta}_{i}$ is the downsampling scale. In this way, each level feature is able to be aware of the specific spatial location with the aid of the equivalent camera intrinsic.


\noindent
\textbf{Hierarchy Depth Prediction.} 
We individually predict the regressed depth $D_{i,j}^{\rm R}\in{\rm{R}}^{H_{i} \times W_{i}}$~\cite{li2023bevdepth} and the categorical depth $D_{i,j}^{\rm C}\in{\rm{R}}^{H_{i} \times W_{i}}$~\cite{wang2022probabilistic} to generate more reliable depth maps. However, in previous methods, depth range $[d_{\rm min}, d_{\rm max} ]$ is divided into multiple fixed-position bins, neglecting the variations among samples and pixels. To handle this issue, we implement the multi-scale pixel-wise refinement of the bin~\cite{bhat2023zoedepth}. Specifically, we first predict the original depth bins $c_{l,j}$ from the highest feature $F_{l,j}'$ at each pixel position with an MLP layer, then $c_{l,j}$ is adjusted to generate another set of depth bins for next layer feature of higher resolution. Take the $i_{th}$ layer feature $F_{i,j}'$ as an example, an MLP layer is introduced to predict $N$ attractor points for each pixel position. The adjusted bin center is $c_{i,j} = c_{i+1,j}+\bigtriangleup c_{i,j}$, this adjustment can be given by:
\begin{equation}
\label{eq6}
\bigtriangleup c_{i,j} = \sum_{n=1}^{N} \frac{p_{n}-c_{i+1,j} }{1+\alpha |p_{n}-c_{i+1,j}|^\beta },
\end{equation}
where $p_{n}$ is the position of $N$ attractor points, $c_{i+1,j}$ is the generated bin center of higher level feature, $\alpha$ and $\beta$ are hyper-parameters. Meanwhile, we predict the probabilistic $P_{i,j}$ over these bins for each pixel, $P\in{\rm{R}}^{N_{\rm B} \times H_{i} \times W_{i}}$, where $N_{\rm B}$ denotes the number of bins for each pixel. Thus, the categorical depth can be formulated as:
\begin{equation}
\label{eq7}
D_{i,j}^{\rm C} = \sum_{k=1}^{N_{\rm B}} P_{i,j}^k\times c_{i,j}.
\end{equation}

Finally, we fuse $D_{i,j}^{\rm R}$ and $D_{i,j}^{\rm C}$ with a hyper-parameters $\omega$, the final depth prediction $D_{i,j}$ can be represented as:
\begin{equation}
\label{eq8}
D_{i,j} = \omega D_{i,j}^{\rm C}+(1-\omega) D_{i,j}^{\rm R}.
\end{equation}

\subsection{Hybrid Depth Supervision}
To address the issue of insufficient supervision for depth estimation due to sparse point clouds, we introduce hybrid depth supervision, which compose of two parts, explicit metric supervision and implicit distribution supervision.

\noindent
\textbf{Explicit Metric Supervision.}
Similar to other methods~\cite{li2023bevdepth, huang2022bevdet4d}, the LiDAR point cloud data is introduced and converted to image view with rotation and translation matrix, then we discard the projected 2.5D points that exceed the resolution of corresponding multi-view features. Consequently, the sparse one-hot pixel-wise depth ground-truth label $D^{gt}$ is acquired and will be the explicit metric supervision for the final depth prediction $D^{pred}$.

\noindent
\textbf{Implicit Distribution Supervision.} We exploit a foundation model, DepthAnything~\cite{yang2024depth}, to generate the pseudo depth label for dense supervision. However, the high generalization ability of DepthAnything comes with the neglect of scale and shift of each sample during multi-dataset joint training~\cite{ranftl2020towards,birkl2023midas}. Accordingly, the pseudo depth maps provide implicit distribution prior instead of exact metric depth values, which are more reliable and general.

To fully leverage the strengths of the foundation model, we exploit the generated relative depth results as pseudo labels $D^{pd}$ for extra supervision of our depth prediction $D_{i,j}$. Specifically, we first take the reciprocal of the predicted depth results $D_{i,j}$, then perform mean-variance normalization~\cite{ali2014data} to generate predicted relative depth results $\widehat{{\frac{1}{D_{i,j}}}}$, which ignore  sample-wise scale and shift variations and can be supervised by pre-generated pseudo labels $D^{pd}$.

\subsection{Positional Depth Encoder and Query Decoder}

\noindent
\textbf{3D Positional Depth Encoder.} For the purpose of obtaining depth-aware 3D features that encompass both semantic features and geometric embeddings, we propose the 3D PDE, similar to 3DPPE~\cite{shu20233dppe}, which introduces point-level positional embedding. Specifically, we generate 3D points based on the predicted depth map $D_{i,j}$ and transform them into LiDAR coordinates to serve as position embeddings. This process can be expressed as
\begin{equation}
\label{eq9}
PE_{j}=Sine(f_{proj}(D_{i,j}, K_j, E_j)) ,
\end{equation}
Where $Sine$ represents the positional embedding function used in DETR~\cite{carion2020end}, and $f_{proj}$ denotes the transformation process from 2D points to 3D space using the camera intrinsics $K_j$ and extrinsics $E_j$. Subsequently, an element-wise addition is performed between the positional embedding $PE_{j}$ and the image features to construct the 3D depth-aware feature set $F_{3D}$.

\noindent
\textbf{Query Decoder.}
We adopt the transformer decoder structure from StreamPETR~\cite{wang2023exploring} to produce the final detection results. Specifically, a fixed number of learnable queries are initially generated and then processed through the propagation decoder, utilizing the 3D features $F_{3D}$ and information from a pre-defined memory queue to facilitate spatial and temporal interactions. Concurrently, with the assistance of motion-aware layer normalization, feature aggregation, and the hybrid attention layer within the propagation transformer, the queries are progressively refined, ultimately yielding the final detection results.

\begin{table*}[t]
\centering
\caption{
Comparison of other methods on the nuScenes \texttt{val} set. ${}^{\dag}$ The backbone benefits from perspective pretraining. The best is in \textbf{bold}.
}
\vspace{-0.2cm}
\resizebox{0.95\linewidth}{!}{
\begin{tabular}{l|c|c|cc|ccccc}
\toprule[1pt]
Method & Backbone & Input Size & NDS$\uparrow$ & mAP$\uparrow$ & mATE$\downarrow$  & mASE$\downarrow$ & mAOE$\downarrow$ & mAVE$\downarrow$ & mAAE$\downarrow$ \\
\midrule
BevDet4D~\cite{huang2022bevdet4d} & ResNet50 & $256 \times 704$ & 45.7 & 32.2 & 0.703 & 0.278 & 0.495 & 0.354 & 0.206 \\
PETRv2~\cite{liu2023petrv2} & ResNet50 & $256 \times 704$ & 45.6 & 34.9 & 0.700 & 0.275 & 0.580 & 0.437 & 0.187 \\
BEVStereo~\cite{li2023bevstereo} & ResNet50 & $256 \times 704$ & 50.0 & 37.2 & 0.598 & 0.270 & 0.438 & 0.367 & 0.190 \\
SOLOFusion~\cite{park2022time} & ResNet50 & $256 \times 704$ & 53.4 & 42.7 & 0.567 & 0.274 & 0.511 & 0.252 & 0.181 \\
\rowcolor{gray!20}
FreqPDE (Ours) & ResNet50 & $256 \times 704$ & \textbf{54.3} & \textbf{43.5} &0.577  &0.270  &0.442  &0.257  &0.199  \\
\midrule
Sparse4Dv2$^{\dag}$~\cite{lin2023sparse4d} & ResNet50 & $256 \times 704$ & 53.8 & 43.9 & 0.598 & 0.270 & 0.475 & 0.282 & 0.179 \\
BEVFormerv2$^{\dag}$~\cite{yang2023bevformer} & ResNet50 & - & 52.9 & 42.3 & 0.618 & 0.273 & 0.413 & 0.333 & 0.188 \\
StreamPETR$^{\dag}$~\cite{wang2023exploring} & ResNet50 & $256 \times 704$ & 55.0 & 45.0 & 0.613 & 0.267 & 0.413 & 0.265 & 0.196 \\
SparseBEV$^{\dag}$~\cite{liu2023sparsebev} & ResNet50 & $256 \times 704$ & 55.8 & 44.8 & 0.581 & 0.271 & 0.373 & 0.247 & 0.190 \\
\rowcolor{gray!20}
FreqPDE (Ours) $^{\dag}$ & ResNet50 & $256 \times 704$ & \textbf{56.2} & \textbf{46.3} &0.575  &0.268  &0.405  &0.245  &0.203  \\
\midrule
StreamPETR$^{\dag}$~\cite{wang2023exploring} & V2-99 & $320 \times 800$ &57.2 &48.2  &0.602  &0.256  &0.372  &0.267  &0.192  \\
\rowcolor{gray!20}
FreqPDE (Ours) $^{\dag}$ & V2-99 & $320 \times 800$ & \textbf{58.6} & \textbf{50.6} & 0.576 & 0.261 & 0.375 & 0.253 & 0.200 \\
\midrule
Far3D$^{\dag}$~\cite{jiang2024far3d} & ResNet101 & $512 \times 1408$ &59.4 &51.0  &0.551  &0.258  &0.372  &0.238  &0.195  \\
Sparse4Dv2$^{\dag}$~\cite{lin2023sparse4d} & ResNet101 & $512 \times 1408$ &59.4 &50.5  &0.548  &0.268  &0.348  &0.239  &0.184  \\
StreamPETR$^{\dag}$~\cite{wang2023exploring} & ResNet101 & $512 \times 1408$ &59.2 &50.4  &0.569  &0.262  &0.315  &0.257  &0.199  \\
\rowcolor{gray!20}
FreqPDE (Ours) $^{\dag}$ & ResNet101 & $512 \times 1408$ & \textbf{60.1} & \textbf{51.9} & 0.562 & 0.258 & 0.324 & 0.242 & 0.196 \\
\bottomrule[1pt]
\end{tabular}
}
\label{tab:val}
\end{table*}

\begin{table*}[t!]
\centering
\caption{
Comparison of other methods on nuScenes \texttt{test} set. These results are reported without test-time augmentation, model ensembling, and any future information. The best is in \textbf{bold}.
}
\vspace{-0.3cm}
\resizebox{0.95\linewidth}{!}{
\begin{tabular}{l|c|c|cc|ccccc}
\toprule[1pt]
Method & Backbone & Input Size & NDS$\uparrow$ & mAP$\uparrow$ & mATE$\downarrow$  & mASE$\downarrow$ & mAOE$\downarrow$ & mAVE$\downarrow$ & mAAE$\downarrow$\\
\midrule
BEVDepth~\cite{li2023bevdepth} & V2-99 & $640\times1600$ & 60.0 & 50.3 & 0.445 & 0.245 & 0.378 & 0.320 & 0.126 \\
CAPE-T~\cite{xiong2023cape} & V2-99 & $640\times1600$ & 61.0 & 52.5 & 0.503 & 0.242 & 0.361 & 0.306 & 0.114 \\
FB-BEV~\cite{liu2023sparsebev} & V2-99 & $640\times1600$ & 62.4 & 53.7 & 0.439 & 0.250 & 0.358 & 0.270 & 0.128 \\
HoP~\cite{zong2023temporal} & V2-99 & $640\times1600$ & 61.2 & 52.8 & 0.491 & 0.242 & 0.332 & 0.343 & 0.109 \\
StreamPETR~\cite{wang2023exploring} & V2-99 & $640\times1600$  & 63.6 & 55.0 & 0.479 & 0.239 & 0.317 & 0.241 & 0.119 \\
SparseBEV~\cite{liu2023sparsebev} & V2-99 & $640\times1600$  & 63.6 & 55.6 & 0.485 & 0.244 & 0.332 & 0.246 & 0.117 \\
Sparse4Dv2~\cite{lin2023sparse4d} & V2-99 & $640\times1600$ & 63.8 & 55.6 & 0.462 & 0.238 & 0.328 & 0.264 & 0.115 \\
\rowcolor{gray!20}
Ours & V2-99 & $640 \times 1600$ & \textbf{64.2} & \textbf{56.0} &0.468  &0.241  &0.315  &0.242  &0.115  \\
\bottomrule[1pt]
\end{tabular}
}
\vspace{-0.2cm}
\label{tab:test}
\end{table*}

\subsection{Detection Head and Loss}
During training stage, we exploit sparse projected LiDAR depth $D^{gt}$ and dense pseudo relative depth $D^{pd}$ simultaneously, where smooth L1 loss~\cite{wang2021fcos3d} and MSE loss~\cite{bishop2006pattern} are used:
\begin{equation}
\label{eq11}
\mathcal{L}_{\rm depth} = \lambda _s\mathcal{L}_s(D_{i,j}, D^{gt})+\lambda _m\mathcal{L}_m(\widehat{{\frac{1}{D_{i,j}}}} , \widehat{{} D^{pd }}),
\end{equation}
where $\lambda _s$ and $\lambda _m$ are the hyper-parameters.
Total Loss given the depth prediction loss $\mathcal{L}_{\rm depth}$, 2D focal sampling loss~\cite{wang2023focal} $\mathcal{L}_{\rm sampl}$ and 3D bounding box regression loss $\mathcal{L}_{\rm reg}$, we adopt the Hungarian Matching to achieve label assignment and the total loss can be formulated as:
\begin{equation}
\label{eq12}
\mathcal{L} = \lambda _1\mathcal{L}_{\rm depth}+\lambda _2\mathcal{L}_{\rm samp}+\lambda _3\mathcal{L}_{\rm reg},
\end{equation}
where $\lambda _1$, $\lambda _2$ and $\lambda _3$ are hyper-parameters to balance the different losses.

\section{Experiments}
\label{sec:Experiment}

\subsection{Dataset and Metrics}
All experiments are conducted on the nuScenes~\cite{caesar2020nuscenes} dataset, which is a widely used public dataset specifically designed for autonomous driving research. The dataset contains six calibrated high-resolution cameras, which cover the surrounding view of the road. It consists of 700 training, 150 validation, and 150 testing sequences, with 34K annotated key frames and 10 object categories. For the 3D detection task, we use nuScenes Detection Score (NDS), mean Average Precision (mAP), and adopt other official True Positive (TP) metrics, including mean Average Translation Error (mATE), mean Average Scale Error (mASE), mean Average Orientation Error (mAOE), mean Average Velocity Error (mAVE), mean Average Attribute Error (mAAE). 


\subsection{Implementation Details}
We set StreamPETR~\cite{wang2023exploring} as the baseline to conduct experiments with ResNet50, ResNet101~\cite{he2016deep}, and VoVNet-99~\cite{wang2021fcos3d} backbones on the nuScenes~\cite{caesar2020nuscenes} dataset without any test-time augmentation or future information. The pretrained weights for ResNet models were obtained from ImageNet-1K~\cite{krizhevsky2012imagenet} and nuImages datasets, while the pretrained weights for VoVNet were provided by DD3D~\cite{park2021pseudo}. For the hierarchy depth prediction, we set $\alpha$ to 300.0, $\beta$ to 2.0, and $\omega$ to 0.5. For the balancing factors of different losses, we set $\lambda _1$, $\lambda _2$, and $\lambda _3$ to 1.0, 0.5 and 1.0.
We adopt image data augmentations, including random cropping, scaling, flipping, and rotation. All models are trained with the streaming video training method~\cite{wang2023exploring} on 4 NVIDIA A100 GPUs with 24 epochs for V2-99 backbone and 90 epochs for ResNet backbone, using AdamW optimizer. The learning rate is set to 4e-4 and batch size is set to 16. Furthermore, all pseudo labels are generated offline and stored as NPZ files, which enable direct loading during training process without additional computational overhead.

\subsection{Main Results}
We compare the proposed FreqPDE with previous state-of-the-art multi-view 3D object detectors on the nuScenes \texttt{val} set. As shown in Tab. ~\ref{tab:val}, our method achieves 56.2\% NDS and 46.3\% mAP performance with the image size of 256 \time 704 and ResNet50 backbone which is pre-trained on nuImages, surpassing other SOTA methods and outperforming the baseline(StreamPETR) by 1.2\% NDS and 1.3\% mAP. When adopting the image size of 320 \time 800 and V2-99 backbone, our method has an obvious performance improvement over the baseline with NDS of 1.4\% and mAP of 2.4\%. Furthermore, under the setting of ResNet101 and high-resolution input, FreqPDE also achieves the highest performance metrics, exceeding the state-of-the-art method (Far3D) by 0.7\% NDS and 0.9\% mAP.
Tab.~\ref{tab:test} presents the performance metrics on the nuScenes \texttt{test} set, in terms of the both metric NDS and mAP, FreqPDE have achieved the SOTA performance, surpassing HoP, StreamPETR and Sparse4dv2.

\subsection{Ablation Study}
\noindent In our ablation experiments, we consistently use V2-99 with nuImage pretraining weights. The input image size is set to 320 × 800.

\noindent
\textbf{Component Analysis.} We conduct ablation studies to study the effectiveness of each proposed module. As shown in Tab.~\ref{component}, compared with the baseline, FSPE brings 0.5 \% NDS performance improvement, indicating the effectiveness of frequency information in detection tasks. And CSDP obtains 1.2 \% mAP improvement, which is non-trivial in promoting the quality of object position regression with the additional depth prediction as an auxiliary task. Moreover, with the help of PDE, our method can achieve consistent 0.4\% NDS performance improvement by exploiting the depth-aware position embedding to construct the 3D features. Combining all designs, FreqPDE achieves a convincing performance upon the baseline, which demonstrates the effectiveness of our proposed method.
\begin{table}[t]\footnotesize
\begin{center}
\caption{Component analysis of FreqPDE on nuScenes \texttt{val} set.} 
\label{component}
\vspace{-0.3cm}
\setlength{\tabcolsep}{0.45cm}
\begin{tabular}{c c c|c|c}
\toprule[1pt]
\textbf{FSPE} & \textbf{CSDP} & \textbf{PDE} & \textbf{NDS $\uparrow$} & \textbf{mAP} $\uparrow$ \\ 
\midrule
\ding{55} & \ding{55}  & \ding{55}& 57.2 & \cellcolor{gray!20}48.2 \\
 \checkmark & \ding{55}  & \ding{55}& 57.7 & \cellcolor{gray!20}48.8 \\
\ding{55} & \checkmark  & \ding{55} & 57.6 & \cellcolor{gray!20}49.4 \\
\ding{55} & \checkmark  & \checkmark &58.0  & \cellcolor{gray!20}49.9 \\
 \checkmark & \checkmark & \checkmark & \textbf{58.6} & \cellcolor{gray!20}\textbf{50.6} \\

\bottomrule[1pt]
\end{tabular}
\end{center}
\vspace{-0.5cm}
\end{table}

\noindent
\textbf{Ablation for design choices in FSPE module.} In our Frequency-aware Spatial Pyramid Encoder, high-frequency and low-frequency information are introduced for different usages, namely boundary enhancement and semantic extraction. As shown in Tab.~\ref{fspe}, introducing the high-frequency information brings significant improvement in mAP metric than NDS, which achieves a 1.0\% increase. With the combination of low-frequency and high-frequency information, our method results in a 1.8\% increase in mAP, reflecting the necessity of enhanced boundary and global semantic for 3D object detection.
\begin{table}[t]\footnotesize
\begin{center}
\caption{Ablation for Frequency-aware Spatial Pyramid Encoder.} 
\label{fspe}
\vspace{-0.2cm}
\setlength{\tabcolsep}{0.3cm}
\begin{tabular}{ c c|c|c}
\toprule[1pt]
 \textbf{Low-Frequency} & \textbf{High-Frequency} & \textbf{NDS $\uparrow$} & \textbf{mAP} $\uparrow$ \\ 
\midrule
\ding{55}  & \ding{55} &57.8   &\cellcolor{gray!20}49.2 \\
 \ding{55}  & \checkmark &57.9  & \cellcolor{gray!20}49.7\\
  \checkmark  & \ding{55}&58.1   &\cellcolor{gray!20}49.6 \\
 \checkmark  & \checkmark & \textbf{58.2}   &\cellcolor{gray!20}\textbf{50.1} \\

\bottomrule[1pt]
\end{tabular}
\end{center}
\vspace{-0.5cm}
\end{table}

\noindent\textbf{Computation Consumption}
We compared the parameter count and inference time of different modules in StreamPETR and FreqPDE, as shown in Table~\ref{consumption}, our method improved NDS by 1.4\% at the cost of a 20MB increase in parameter size, primarily due to the multi-layer structures of FSPE and CSDP. To address this issue, we also evaluated a single-layer variant, which reduced the parameter increase to 6MB while still achieving a 0.9\% NDS improvement, with FPS remaining nearly unchanged. This highlights the efficiency and resource effectiveness of our approach. 
\begin{table}[t]\footnotesize
\begin{center}
\caption{Ablation of Computation Consumption.} 
\label{consumption}
\vspace{-0.2cm}
\setlength{\tabcolsep}{0.3cm}
\resizebox{\columnwidth}{!}{
\begin{tabular}{c|ccc|c|c|c}
\toprule[1pt]
Method                      & Backbone & Neck  & Head   & Total   & NDS                   & FPS                   \\
\midrule
\multirow{2}{*}{StreamPETR} & 66.3MB   & 1.0MB & 11.8MB & 79.1MB  & \multirow{2}{*}{57.2} & \multirow{2}{*}{\textbf{10.7}} \\
                            & 20.1ms   & 0.8ms & 71.3ms & 92.2ms  &                       &                       \\
\midrule                   
\multirow{2}{*}{FreqPDE}    & 66.3MB   & 4.1MB & 29.9MB & 100.3MB & \textbf{58.6} & \multirow{2}{*}{8.8}  \\
                            & 20.1ms   & 3.5ms & 88.8ms & 112.4ms & \textcolor{red}{(+1.4)}                      &                       \\
\midrule
\multirow{2}{*}{FreqPDE-L1} & 66.3MB   & 2.1MB & 16.6MB & 85.0MB  & 58.1  & \multirow{2}{*}{10.2} \\
                            & 20.1ms   & 1.6ms & 76.7ms & 98.4ms  & \textcolor{red}{(+0.9)}                      &                       \\
\bottomrule[1pt]
\end{tabular}}
\end{center}
\vspace{-0.5cm}
\end{table}

\noindent
\textbf{Ablation for design choices in CSDP module.} We compare various designs for predicting the depth map in Tab.~\ref{csdp}. The hierarchical depth prediction approach boost the performance with a 2.0\% increase in mAP, highlighting the necessity of depth prediction for addressing scale invariance. Furthermore, incorporating efficient channel attention and cross-view attention leads to notable improvements of 1.8\% mAP and 0.9\% NDS, demonstrating the effectiveness of multi-scale camera parameter embeddings and adjacent perspective interactions.
\begin{table}[t]\footnotesize
\begin{center}
\caption{Ablation for Cross-view Scale-invariant Depth
Predictor.} 
\label{csdp}
\vspace{-0.3cm}
\setlength{\tabcolsep}{0.45cm}
\begin{tabular}{c c c|c|c}
\toprule[1pt]
\textbf{HDP} &\textbf{ECA} & \textbf{CVA}  & \textbf{NDS $\uparrow$} & \textbf{mAP} $\uparrow$ \\ 
\midrule
 \ding{55} & \ding{55}  & \ding{55}&57.7  & \cellcolor{gray!20}48.8 \\
\checkmark & \ding{55}  & \ding{55}&58.1  & \cellcolor{gray!20}49.8 \\
 \checkmark & \checkmark  & \ding{55}&58.2  & \cellcolor{gray!20}50.0 \\
\checkmark & \checkmark  & \checkmark & \textbf{58.6}  &\cellcolor{gray!20}\textbf{50.6} \\

\bottomrule[1pt]
\end{tabular}
\end{center}
\vspace{-0.6cm}
\end{table}

\noindent
\textbf{Pseudo Depth Label Supervision.} 
To validate the effectiveness of Implicit Distribution Supervision, we analyze detection performance at different distances, as shown in Table~\ref{new_range}. As the distance increases, the method that combines sparse LiDAR supervision with dense pseudo supervision achieves increasingly significant improvements in NDS. Specifically, dense pseudo labels bring an improvement of 1.9\% , while sparse LiDAR labels only bring an improvement of 0.6\% when the distance exceeds 40 meters. Our results demonstrate that pseudo labels notably enhance long-range object detection, reinforcing our approach to mitigating sparse supervision in distant regions. 

\begin{table}[t]\footnotesize
\begin{center}
\caption{
Comparison in different distances. NDS$_{\textgreater0}$, NDS$_{\textgreater20}$, and NDS$_{\textgreater40}$ represent different evaluation metrics under distance thresholds of 0, 20, and 40 meters, respectively.} 
\label{new_range}
\vspace{-0.2cm}
\setlength{\tabcolsep}{0.2cm}
\resizebox{1.0\columnwidth}{!}{
\begin{tabular}{c|c|c|c|c}
\toprule[1pt]
\textbf{Method}  &\textbf{Supervision} & \textbf{NDS$_{\textgreater 0}$} $\uparrow$ & \textbf{NDS$_{\textgreater 20}$} $\uparrow$& \textbf{NDS$_{\textgreater 40}$} $\uparrow$ \\ 
\midrule
 Baseline   &- &57.2  &38.2  &13.3  \\
 FreqPDE &Sparse &58.4 \textcolor{red}{(+1.2)}  &39.3 \textcolor{red}{(+1.1)}  &13.9 \textcolor{red}{(+0.6)} \\
 FreqPDE &Sparse + Dense &\textbf{58.6} \textcolor{red}{(+1.4)} &\textbf{39.7} \textcolor{red}{(+1.5)}  &\textbf{15.2} \textcolor{red}{(+1.9)}  \\
\bottomrule[1pt]
\end{tabular}
}
\end{center}
\vspace{-0.2cm}
\end{table}

\subsection{Qualitative Results} 
\begin{figure}[t]
\centering
\includegraphics[width=0.9\columnwidth]{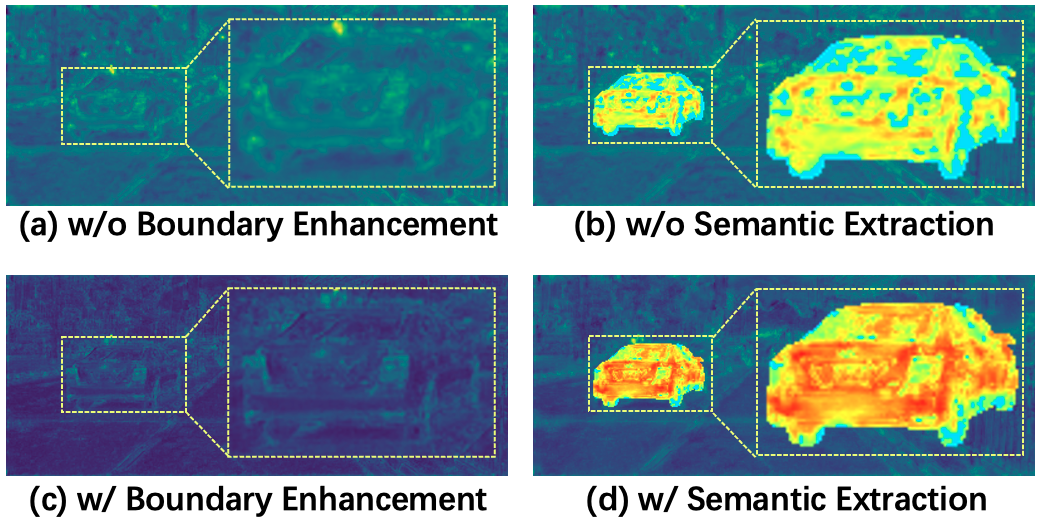}
\vspace{-0.3cm}
\caption{Feature visualization and intra-category similarity (IntraSim). The brighter color indicates a higher IntraSim for the car. }
\label{freq_vis}
\vspace{-0.2cm}
\end{figure}

To examine the quality of extracted features from the proposed FSPE module, we visualize the image features before and after using the corresponding module. As shown in Fig.~\ref{freq_vis} (a) and (c), with the helpf of high-frequency boundary enhancement module, the feature contours are more distinct, with more consistent background. We also calculate the intra-category similarity~\cite{chen2024frequency} where the brighter color shows the high similarity within the same car category, indicating that the feature Fig.~\ref{freq_vis} (d) obtains more global semantic information than Fig.~\ref{freq_vis} (b), thanks to the utilization of low-frequency semantic extraction module.
\section{Conclusion}
\label{sec:Conclusion}
In this paper, we propose the FreqPDE, a novel positional depth embedding method for multi-view 3D object detection transformers. Different from the previous methods, our work mainly harnesses the frequency-dimensional information across multi-level features which are encoded to generate the 3D depth-aware features for complementary perception. Specifically, we first utilize FSPE to construct frequency-aware 2D features. Then, CSDP is introduced to predict the pixel-wise depth map with a hierarchy predictor. Besides, the positional depth encoder generated the 3D depth-aware features which are fed into the transformer decoder to output the final detection results. Furthermore, sufficient experiments have been conducted on the nuScenes dataset to validate the effectiveness and feasibility of our proposed components. Finally, we hope FreqPDE could promote the research of frequency domain and depth in multi-view 3D object detection.
\section{Acknowledgments}
\label{sec:ack}

This paper is in part supported by National Natural Science Foundation of China (No. 62088102) and Shanghai Municipal Science and Technology Major Project, China (No. 2021SHZDZX0102).

{
    \small
    \bibliographystyle{ieeenat_fullname}
    \bibliography{main}
}
\clearpage
\setcounter{page}{1}
\maketitlesupplementary

\appendix
\section{Appendix}
\subsection{More Related Work}
\textbf{Depth Estimation}
Depth estimation from 2D camera images is a challenging topic in Computer Vision, categorized into regressing metric depth~\cite{bhat2021adabins,yuan2022new,jun2022depth,bhat2022localbins,li2024binsformer} and relative depth~\cite{mertan2022single,lee2019monocular,ranftl2021vision}. BinsFormer~\cite{li2024binsformer} introduces sufficient interaction between probability distribution and bin predictions to generate proper metric estimation.  DPT~\cite{ranftl2021vision} exploits vision transformers as a backbone for dense relative depth prediction. Recent works~\cite{bhat2023zoedepth,yang2024depth,yang2024depth2} attempt to build a foundation model with excellent generalization performance across domains while maintaining metric scale. ZoeDepth~\cite{bhat2023zoedepth} uses a lightweight depth head with a novel metric bin design to combine metric and relative depth estimation. DepthAnything~\cite{yang2024depth,yang2024depth2} introduces the affine-invariant loss to ignore the unknown scale and shift during the training stage, additionally, a data engine has been devised to automatically generate pseudo depth annotations for unlabeled images.

\noindent
\textbf{3D Positional Embedding}
The necessity of the 3D Position Encoder (PE) has been addressed in prior studies~\cite{liu2022petr, liu2023petrv2, shu20233dppe}. PETR series~\cite{liu2022petr, liu2023petrv2} discretize the camera frustum space into meshgrid coordinates which are transformed to 3D world space with camera parameters, then the 3D coordinates are input to a 3D position encoder with 2D image features to construct the 3D position-aware features. However, leveraging hand-crafted camera-ray depth bins as the channel dimensionality for the point cloud disregards depth variations across different pixels. To ameliorate the aforementioned problem, 3DPPE~\cite{shu20233dppe} transforms the pixels to 3D space with camera parameters and predicted pixel-wise depth results, the resulting 3D points are sent to a position encoder to construct the 3D feature with point-level embeddings.

\subsection{Implicit Distribution Supervision.} 
To fully leverage the strengths of the foundation model, we exploit the generated relative depth results as pseudo labels for extra supervision of our depth prediction $D_{i,j}$. To elaborate, the crucial issue is converting metric depth to relative depth, this process can be formulated as:
\begin{equation}
\label{eq9}
D^{rel} = \frac{1}{{\rm scale}}(\frac{1}{D^{mtr}}-{\rm shift} ) ,
\end{equation}
where $D^{rel}$ is relative depth, $D^{mtr}$ is metric depth, $\rm scale$ and $\rm shift$ are sample-wise parameters for transposition. Noticing the linear relationship between $\frac{1}{D^{mtr}}$ and $D^{rel}$, we perform mean-variance normalization~\cite{ali2014data} separately:
\begin{equation}
\label{eq10}
\begin{aligned}
&\widehat{\frac{1}{D^{mtr}}} = \frac{\frac{1}{D^{mtr}}-\mathbb{E}(\frac{1}{D^{mtr}}) }{\sqrt{Var(\frac{1}{D^{mtr}})} }, \\&\widehat {{D^{rel}}} = \frac{\frac{1}{\rm scale}(\frac{1}{D^{mtr}}-\mathbb{E}(\frac{1}{D^{mtr}})) }{|\frac{1}{\rm scale}|\sqrt{Var(\frac{1}{D^{mtr}})} },
\end{aligned}
\end{equation}
where $\mathbb{E}$ and ${Var}$ represent the computation of the mean and variance respectively, $\widehat{\frac{1}{D^{mtr}}}$ and $\widehat{D^{rel}}$ correspond to the normalized outcomes. Given that the coefficient $\frac{1}{\text{scale}}$ is strictly positive, it follows that the two normalization results for each sample are equivalent. Consequently, we take the reciprocal of the predicted depth $D_{i,j}$, normalize the outcome, and employ the normalized pseudo-labels as supervisory signals to facilitate the supervised learning transition from metric depth to relative depth.

\subsection{More Ablation Study}
\noindent
\textbf{Cross View Attention.}
In our CSDP module, we apply a fixed-ratio mask to the features in order to mitigate the influence of non-overlapping regions. To verify the effectiveness of this masking approach, we conduct ablation studies to evaluate the impact of different mask ratios, as illustrated in Tab.~\ref{cross_view}. With a mask ratio of 0.2, our method demonstrates improved performance, outperforming the model without masking and other mask ratio.
\begin{table}[h] \footnotesize
\begin{center}
\caption{Necessity of cross-view.} 
\label{cross_view}
\vspace{-0.3cm}
\setlength{\tabcolsep}{0.2cm}
\begin{tabular}{c|c|c|c}
\toprule[1pt]
\textbf{Mask Ratio}  & \textbf{NDS $\uparrow$} & \textbf{mAP} $\uparrow$ & \textbf{mATE}$\downarrow$\\ 
\midrule
 -  & 58.3 & 50.3 &0.578\\
0.1   & 57.3 & 49.6  &0.609\\
 0.2  & \textbf{58.5} & \textbf{50.5} &\textbf{0.569}\\
 0.3  & 58.2 & 50 &0.580\\

\bottomrule[1pt]
\end{tabular}
\end{center}
\vspace{-0.2cm}
\end{table}

\noindent
\textbf{Effect of Positional Depth Encoder} This study seeks to provide empirical evidence of that positional encoding, within the multi-level depth maps, enhances the detection capacity of 3D objects by the query. As shown in Tab.~\ref{pde}, wherein multi-level scale-invariant depth prediction results are resized to the same scale and fused together to be fed into a point-wise embedding function, which outperforms the baseline by 1.4\% NDS and 2.4\% mAP, also exceed the single-level embedding method similar to 3DPPE~\cite{shu20233dppe}.
\begin{table}[t]\footnotesize
\begin{center}
\caption{Ablation for Positional Depth Encoder on nuScenes.} 
\label{pde}
\vspace{-0.3cm}
\setlength{\tabcolsep}{0.5cm}
\begin{tabular}{c|c|c|c}
\toprule[1pt]
\textbf{Method} & \textbf{NDS $\uparrow$} & \textbf{mAP} $\uparrow$ & \textbf{mATE} $\downarrow$  \\ 
\midrule
 Baseline  & 57.2 & \cellcolor{gray!20}48.2 & 0.602 \\
 Single-level  & 57.9 & \cellcolor{gray!20}49.6 & 0.587 \\
 Multi-levels  & \textbf{58.6} & \cellcolor{gray!20}\textbf{50.6} & \textbf{0.576} \\

\bottomrule[1pt]
\end{tabular}
\end{center}
\vspace{-0.4cm}
\end{table}

\noindent\textbf{Comparison with LSS method}
To validate the 'plug-and-play' capability of proposed depth predictor, we replace the depth predictor in BEVDepth with our FSPE and CSDP modules. The results, presented in Table~\ref{lss}, demonstrate the effectiveness and transferability of our proposed design.
\begin{table}[t]\footnotesize
\begin{center}
\caption{Comparison with LSS-based method.} 
\label{lss}
\vspace{-0.3cm}
\setlength{\tabcolsep}{0.2cm}
\resizebox{1.0\columnwidth}{!}{
\begin{tabular}{c|c|c|c|c}
\toprule[1pt]
Method      & Backbone & Input Resolution & mAP  & NDS  \\

\midrule
BEVDepth     & R50    & 256*704         & 35.1 & 47.5 \\

BEVDepth-R        & R50    & 256*704         & \textbf{36.0} & \textbf{48.4}  \\

\bottomrule[1pt]
\end{tabular}}
\end{center}
\vspace{-0.2cm}
\end{table}

\noindent
\textbf{Effect of Hybrid Depth Supervision.} To further validate the effectiveness of the hybrid supervision approach for CSDP, we compare the performance of different supervision methodologies. As presented in Tab.~\ref{hybrid_super}, employing only pseudo-labels results in an improvement in detection performance; however, it leads to a decrease in depth estimation performance. This indicates that distribution-based supervision provides a more comprehensive supervisory signal for overall depth maps but lacks the precision of absolute depth supervision. Consequently, with hybrid supervision, both the absolute relative error (Abs Rel) and squared relative error (Sq Rel) decrease, while the model achieves a 1.4\% increase in mAP and a 0.6\% increase in NDS.
\begin{table}[t]\footnotesize
\begin{center}
\caption{Effect of Hybrid Depth Supervision on nuScenes \texttt{val} set.} 
\label{hybrid_super}
\vspace{-0.3cm}
\setlength{\tabcolsep}{0.3cm}
\begin{tabular}{c|c|c|c|c}
\toprule[1pt]
\textbf{Supervision} & \textbf{Abs Rel} $\downarrow$ & \textbf{Sq Rel} $\downarrow$ & \textbf{NDS $\uparrow$} & \textbf{mAP} $\uparrow$ \\ 
\midrule
 LiDAR only & 0.17 &1.45  & 58.3 & \cellcolor{gray!20}49.9 \\
 Pseudo only & 0.23 &3.71  & 58.4 & \cellcolor{gray!20}50.1 \\
 Hybrid & \textbf{0.15} & \textbf{1.41} & \textbf{58.6} & \cellcolor{gray!20}\textbf{50.6} \\

\bottomrule[1pt]
\end{tabular}
\end{center}
\vspace{-0.2cm}
\end{table}

\subsection{Result Visualization}
\begin{figure*}[t]
\centering
\includegraphics[width=\textwidth]{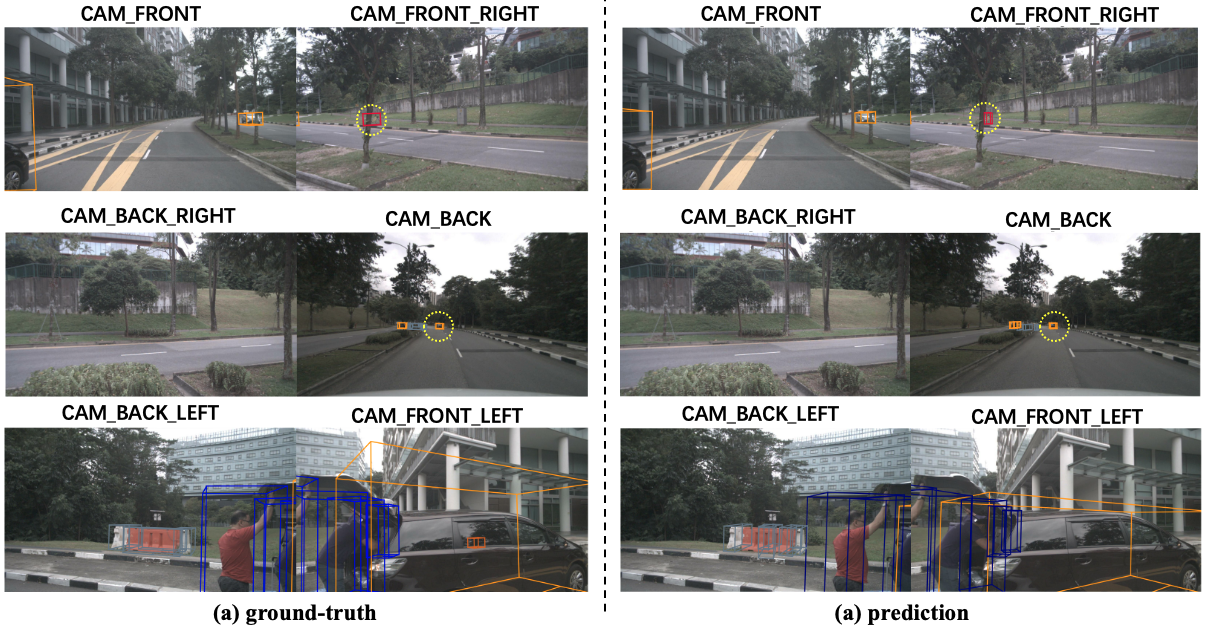}
\vspace{-0.4cm}
\caption{Qualitative detection results on multi-view images on the
nuScenes val set. The 3D predicted bounding boxes are drawn with different colors for different classes.}
\label{fig_vis}
\vspace{-0.2cm}
\end{figure*}

\noindent
\textbf{Qualitative Results.} We show the qualitative detection results of FreqPDE in Fig.~\ref{fig_vis} on multi-view images. The 3D predicted bounding boxes are drawn with different colors for different classes. As illustrated by the highlighted circles, our method accurately detects the category and location of distant targets, while also mitigating the challenges posed by occluded small targets to some extent. This demonstrates an enhancement in the model's detection capability for distant targets following the integration of a more precise depth estimation module.

\begin{figure*}[t]
\centering
\includegraphics[width=\textwidth]{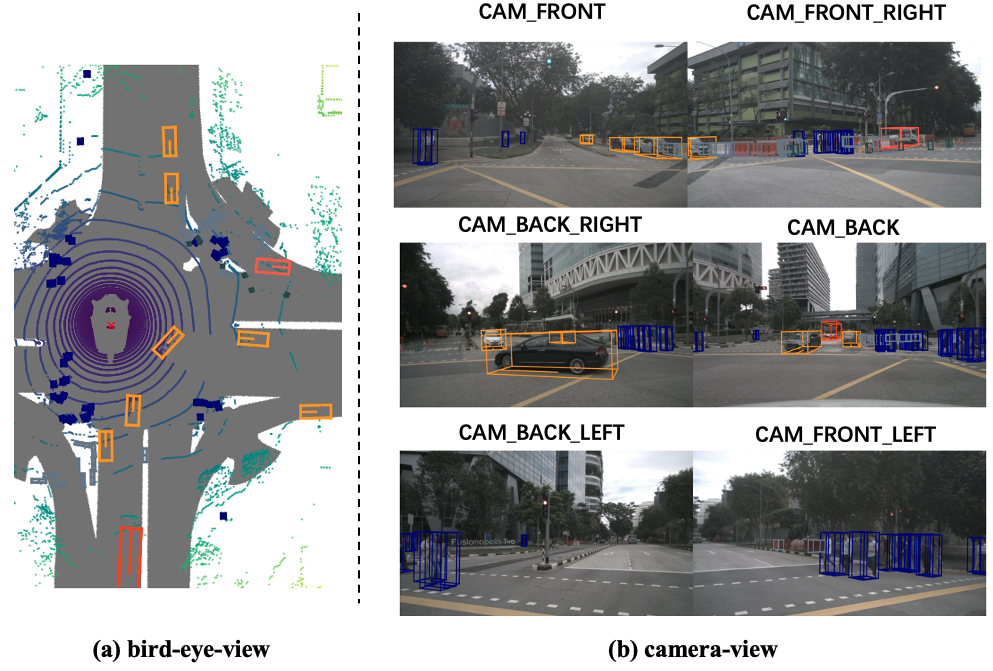}
\vspace{-0.4cm}
\caption{Qualitative detection results on multi-view images and BEV space on the nuScenes val set.}
\label{fig_vis1}
\vspace{-0.2cm}
\end{figure*}

\begin{figure*}[t]
\centering
\includegraphics[width=\textwidth]{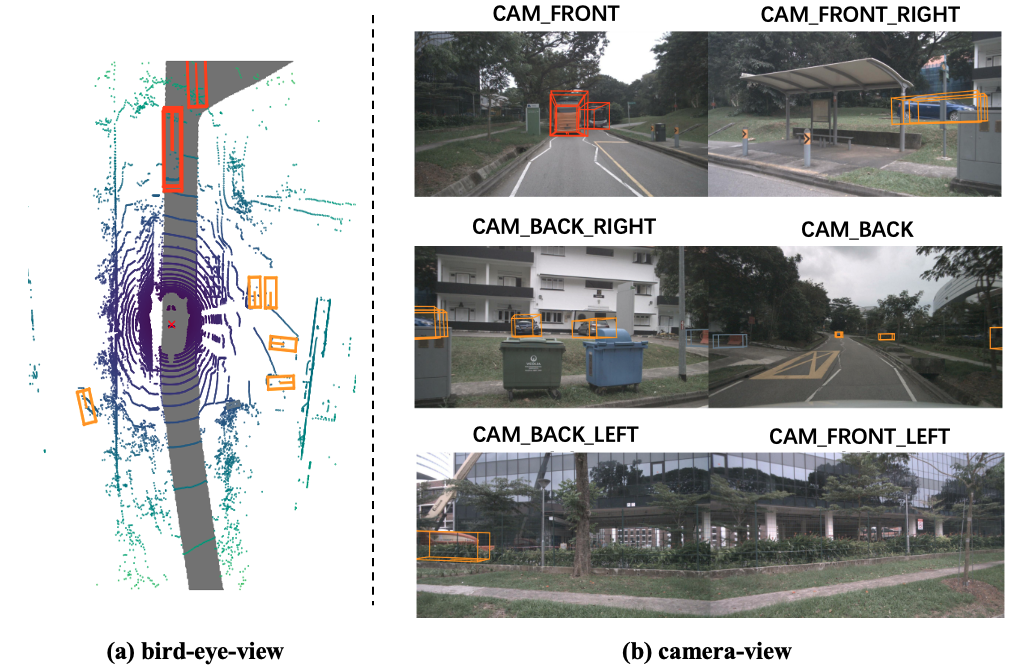}
\vspace{-0.4cm}
\caption{Qualitative detection results on multi-view images and BEV space on the nuScenes val set.}
\label{fig_vis2}
\vspace{-0.2cm}
\end{figure*}

\noindent
\textbf{More Visualization.} We also show more detection results of some challenging scenes in Fig.~\ref{fig_vis1} and Fig.~\ref{fig_vis2}. Our method shows impressive results on crowded and distant objects.

\end{document}